\documentclass[letter, 10pt, conference]{ieeeconf}
\IEEEoverridecommandlockouts

\usepackage{cite}

\usepackage{amsmath,amssymb,amsfonts}
\usepackage{algorithmic}
\usepackage{graphicx}
\usepackage{subcaption}
\usepackage{textcomp}
\usepackage{xcolor}
\usepackage{bm}
\usepackage{booktabs}
\usepackage{svg}
\svgsetup{inkscapelatex=false}
\usepackage{tikz}
\usepackage{hyperref}

\def\BibTeX{{\rm B\kern-.05em{\sc i\kern-.025em b}\kern-.08em
    T\kern-.1667em\lower.7ex\hbox{E}\kern-.125emX}}
\usetikzlibrary{arrows,decorations.pathmorphing,backgrounds,fit,positioning,calc,shapes}

\begin{document}

\title{Behavior Trees for Robust Task Level Control in Robotic Applications
\thanks{This project is financially supported by the Swedish Foundation for Strategic Research.}
}

\author{\authorblockN{
Matteo Iovino$^{a,b}$,
Christian Smith$^{a}$}
\thanks{$^{a}$Division of Robotics, Perception and Learning, KTH - Royal Institute of Technology, Stockholm, Sweden}
\thanks{$^{b}$ABB Corporate Research, Västerås, Sweden}
}

\maketitle

\begin{abstract}
Behavior Trees are a task switching policy representation that can grant reactiveness and fault tolerance. Moreover, because of their structure and modularity, a variety of methods can be used to generate them automatically. In this short paper we introduce Behavior Trees in the context of robotic applications, with overview of autonomous synthesis methods.
\end{abstract}


\section{Introduction}
In the past decades the number and variety of applications that have been robotised has increased drastically. Historically, robots have been used in the manufacture industry to increase the production time and scale. Here, robots operate in a cell and perform the same repetitive task because the environment they act on is deterministic. In new applications, robots are deployed in unpredictable environments instead, where they have to account for random events, e.g. moving obstacles, interaction with humans, uneven terrains. If the robot is more prone to fail, to ensure a successful mission under these circumstances, it is necessary to control it with a reactive and fault tolerant policy. 
Behavior Trees (BTs) are a good choice to fulfill these requirements because of their reactivity, i.e. the ability to react to unforeseen events, modularity, i.e. the possibility to easily edit the policy and reuse part of it and explainability, i.e. how easy it is to an human operator to understand the behavior of the robot at runtime~\cite{colledanchise_behavior_2018}. The interest of the robotic research community on Behavior Trees is increasing, spanning their usage over a large set of applications and defining new methods for their automatic synthesis, to reduce the effort in terms of time and programming skills that is required to generate them.
In this paper we will define BTs in the context of a robotic application and we will present some available methods to synthesise them automatically, starting from a set of robot skills.

\begin{figure}[tbp]
    \centering
    \includegraphics[width=\linewidth]{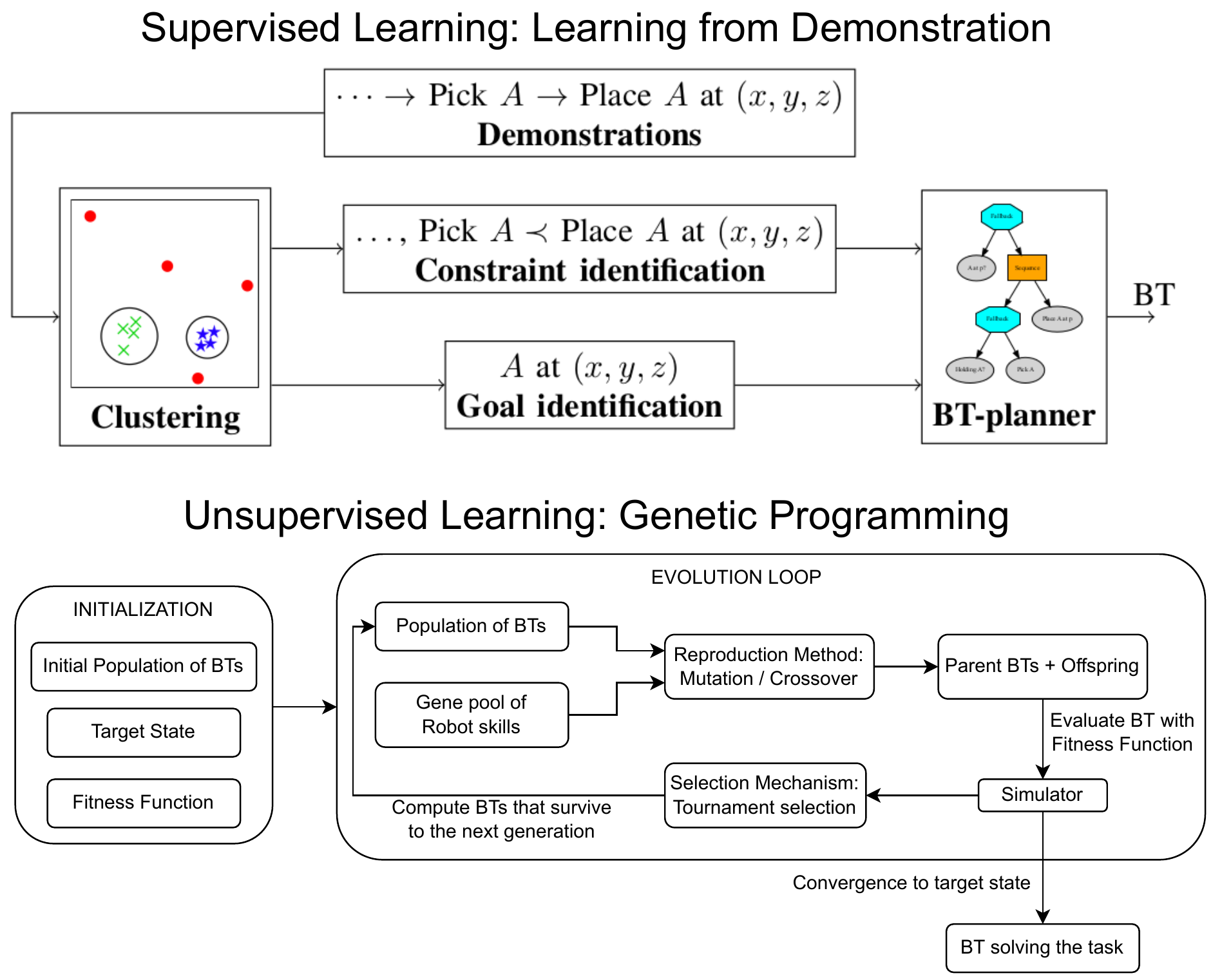}
    \caption{Functional scheme of the proposed learning methods for intuitive and efficient robot behavior creation for high-level control.}
    \label{fig:learning_BT}
\end{figure}

\section{Behavior Trees}

Behavior Trees (BTs) are task switching policy representations that originated in the gaming industry as an alternative architecture to Finite State Machines (FSM) for the control of artificial agents~\cite{colledanchise_behavior_2018}. They have explicit support for task hierarchy, action sequencing and reactivity, and improve on FSMs especially in terms of modularity and reusability~\cite{iovino_survey_2022}.
In their standard formulation, BTs feature three types of internal nodes (called \emph{control nodes}), that define the execution policy of the children, and two types of leaves or terminals (called \emph{execution nodes}), that encode robot skills and status checks. The execution is enabled by a tick signal that originates from the root of the tree at a given frequency and that propagates down the tree from left to right. Every node executes when it receives a tick and returns one of the status signals \textit{Running}, \textit{Success} or \textit{Failure}. For more details on BTs, please refer to~\cite{colledanchise_behavior_2018, iovino_survey_2022}.
BTs can functionally be compared to decision trees, but the \textit{Running} state allows BTs to execute actions for longer than one tick. The \textit{Running} state is also crucial for reactivity, allowing other actions to preempt non-finished ones. Modularity is achieved with the standardized I/O (tick/return statuses) structure of all behaviors~\cite{biggar_modularity_2022}.

\section{Automatic Synthesis of Behavior Trees}

Autonomous generation of BTs can be realized both with planning and learning methods~\cite{iovino_survey_2022} or a combination of those~\cite{styrud_combining_2022}. In this short paper we provide some examples of methods for BT learning in robotic applications, specifying the motivations to use them. A schematic outline is reported in Fig.~\ref{fig:learning_BT}.

\paragraph{Planning}
Planning definition languages, such as PDDL, can be used to define the knowledge base for the robot. In such definitions, the skills are defined together with their pre- and post-conditions as well as the parameters. With such representation, it is possible to use planners to automatically generate BTs. Starting from the goal condition for the task (e.g. `Item Delivered?'), actions that achieve it are chosen (e.g. `Place Item!'), i.e. those actions that have that particular condition as their post-condition. Then, those actions' unmet pre-conditions (e.g. `Item in Hand?' and `Robot-At Delivery Station?' ) are expanded in the same way. This solution strategy allows to generate the so-called backward-chained BTs. Autonomous planners have the advantage of the short computation time, but their success depends on the amount of domain knowledge they are provided. In~\cite{styrud_combining_2022}, we use planned BTs to bootstrap a learning BTs with Genetic Programming (GP) for assembly tasks where different LEGO DUPLO bricks have to be combined in some configurations. If the task is simple enough, like stacking three bricks, the planner alone finds a solution. If however the planner is missing some important information about the environment the robot operates in, like physics and balance, then it fails, as we showed in a task where the robot has to elevate a large brick with uneven balance.

\paragraph{Learning from Demonstration}
Learning from Demonstration (LfD) is a method to create robot programs that leverage the experience of the users. A task is shown to the robot by means of kinesthetic teaching, teleoperation or passive observation. We generate BTs automatically from demonstrations of manipulation tasks~\cite{gustavsson_combining_2022, iovino_interactive_2022} (Fig.~\ref{fig:learning_BT}, top). The learnt policy is robust to initial configurations of the pose of the items necessary to complete the task. Moreover, if the item to manipulate is ambiguously defined, the robot will ask disambiguation question to the human to identify the target for the task.
LfD methods are intuitive and require little programming knowledge, but the more complicated the task, the more cumbersome it becomes to instruct the robot. With our method~\cite{gustavsson_combining_2022}, we are able to solve kitting-like problems, where the robot has to prepare a kit box for a human operator, or assembly-like problems, where the robot has to perform a specific sequence of actions to combine small components. With three demonstration, the robot is able to solve the task with any starting configuration and react to external disturbances. For example, in a stacking task of three cubes, if the operator places the first two, then the robot will complete the task by placing the third and if the operator removes one or two cubes after the task is completed, then the robot will replace them again. The method has good performances if the human does not deliberately try to trick the robot during the demonstrations. Our work~\cite{iovino_interactive_2022} opens up to multimodal interaction modalities, where kinesthetic teaching and speech recognition can be used to solve manipulation tasks in a collaborative scenario.


\paragraph{Genetic Programming}
Genetic Programming (GP) is an evolutionary algorithm designed to learn computer programs in an unsupervised fashion. This method is a good fit to learn BTs due to the tree structure and the modularity that allows the GP to easily modify the BT with operations of mutation (adding, deleting or changing a node) and crossover (subtree swap among parents)~\cite{iovino_learning_2021} (Fig.~\ref{fig:learning_BT}, bottom). A fitness function has to be engineered to allow the method to evaluate a population of BTs.
GP is a powerful optimization method but it requires many learning episodes to converge. It is however possible to bootstrap the learning with examples originating from planned solutions~\cite{styrud_combining_2022} or LfD. Since the goal of the method is to learn high-level controllers for the task, the curse of dimensionality can be mitigated by using a fast simulator that does not consider the dynamics of the robot and the objects nor the physics of the environment. With the method proposed in~\cite{iovino_learning_2021}, we learn a BT for a mobile manipulation task in a simplified simulator, where we add probabilistic transitions between states and failure states in case of unsuccessful actions. Then, we validate the learnt tree in a simulator that includes physics where the actions can fail due to localization errors during motion or slippage during picking. The fitness function is designed to take into account the goal configuration of the item to fetch, the overall execution time and the size of the final BT.
The BT is robust to those failure cases that were considered during the learning phase. In~\cite{styrud_combining_2022} we show that bootstrapping the GP with planned solutions reduces the time required for convergence.

\section{Conclusion}

Autonomous synthesis methods haven been used by the robotics community to easily and time efficiently generate and modify fault tolerant task switching controllers. We believe that BTs are a good choice for the controller of robots operating in unpredictable environments. Modularity is a key property for many of these methods because it makes edit operations on the BTs not dependent on its size~\cite{iovino_programming_2022}. We are able to automatically generate BTs for item fetching and delivery, kitting and assembly, because we targeted use cases for the manufacturing industry. However, the methods are generalizable to other tasks as well, provided that it is possible to define a fitness function or a knowledge base for the robot.

\bibliographystyle{IEEEtran}
\bibliography{references.bib}

\end{document}